\documentclass[journal]{IEEEtran}

\usepackage{url}
\usepackage{tikz,pgfplots,pgflibraryplotmarks}
\usepackage{graphicx}
\usepackage{subfig}
\usepackage{mathrsfs}
\usepackage{tcolorbox}
\usepackage{amssymb,amsfonts,amsmath,latexsym,dsfont}
\usepackage{adjustbox}
\usepackage{subfig}
\usepackage{algpseudocode}
\usepackage{algorithm}
\usepackage[noadjust]{cite}

\usepackage[normalem]{ulem}

\ifCLASSINFOpdf
\else
\fi

\newcommand{\bfC}{{\bf C}}

\newcommand{\bfF}{{\bf F}}

\newcommand{\bfI}{{\bf I}}

\newcommand{\bfK}{{\bf K}}

\newcommand{\bfP}{{\bf P}}

\newcommand{\bfW}{{\bf W}}
\newcommand{\bfX}{{\bf X}}
\newcommand{\bfY}{{\bf Y}}

\newcommand{\bfc}{{\bf c}}

\newcommand{\bfy}{ {\bf y}}

\newcommand{\bftheta}{{\boldsymbol \theta}}

\newcommand{\bfmu}{{\boldsymbol \mu}}

\newcommand{\cin}{c_{\rm in}}
\newcommand{\cout}{c_{\rm out}}
\newcommand{\nTheta}{p}
\newcommand{\nf}{n_{\rm f}}
\newcommand{\nc}{n_{\rm c}}
\newcommand{\nOut}{n_{\rm out}}
\newcommand{\R}{\ensuremath{\mathds{R}}}

\begin{document}
\title{LeanConvNets: Low-cost Yet Effective Convolutional Neural Networks}

\author{Jonathan Ephrath, Moshe Eliasof, Lars Ruthotto, Eldad Haber and Eran Treister \thanks{J. Ephrath, M. Eliasof, and E. Treister are with the Department of Computer Sciences at the Ben-Gurion University of the Negev, Be'er Sheva, Israel. emails: {\tt[ephrathj,eliasof]@post.bgu.ac.il, erant@cs.bgu.ac.il.} JE and ME contributed equally to this work.}
\thanks{L. Ruthotto is with the Departments of Mathematics and Computer Science, Emory University, Atlanta, GA, USA. email: \tt{lruthotto@emory.edu}.}
\thanks{E. Haber is with the Department of Earth, Ocean and Atmospheric Sciences, University of British Columbia, and with Xtract AI, Vancouver, Canada. email: {\tt ehaber@eos.ubc.ca}.}
}

%

\maketitle

\begin{abstract}
Convolutional Neural Networks (CNNs) have become indispensable for solving machine learning tasks in speech recognition, computer vision, and other areas that involve high-dimensional data.
A CNN filters the input feature using a network containing spatial convolution operators with compactly supported stencils.
In practice, the input data and the hidden features consist of a large number of channels, which in most CNNs are fully coupled by the convolution operators.
This coupling leads to immense computational cost in the training and prediction phase.
In this paper, we introduce LeanConvNets that are derived by sparsifying fully-coupled operators in existing CNNs.
Our goal is to improve the efficiency of CNNs by reducing the number of weights, floating point operations and latency times, with minimal loss of accuracy.
Our lean convolution operators involve tuning parameters that controls the trade-off between the network's accuracy and computational costs. These convolutions can be used in a wide range of existing networks, and we exemplify their use in residual networks (ResNets).
Using a range of benchmark problems from image classification and semantic segmentation, we demonstrate that the resulting LeanConvNet's accuracy is close to state-of-the-art networks while being computationally less expensive. In our tests, the lean versions of ResNet in most cases outperform comparable reduced architectures such as MobileNets and ShuffleNets.
\end{abstract}


\IEEEpeerreviewmaketitle

\section{Introduction}

\IEEEPARstart{C}{onvolutional} neural networks (CNNs)~\cite{LeCun1990} are among the most effective machine learning approaches for processing structured, high-dimensional data such as voice recordings, images, and videos and have become indispensable in,  e.g., speech recognition \cite{RainaEtAl2009,speechJSTSP}, audio processing~\cite{audioJSTSP}, and image classification~\cite{KrizhevskySutskeverHinton2012}.

In the forward propagation, a CNN filters the input features through a sequence of layers, which are composed of convolution operators, biases, normalization layers, nonlinear activation functions, and pooling operators.
In imaging tasks, the input features and the hidden features at each layer can be grouped into several channels, each of which can be interpreted as an image.
The stencils that parameterize the convolution operators are typically chosen to have a small support around the origin.
Hence, each feature in an image interacts with features from a small neighborhood in its channel and, in the standard, \emph{fully-coupled} approach, the features from the same neighborhood in the remaining channels~\cite{Gu:2018id,Goodfellow:2016wc}.
A drawback of the fully-coupled approach is that the number of convolution operators in a layer is proportional to the product of the number of input and output channels.
This scaling can render wide architectures (i.e., architectures whose layers contain a large number of channels) prohibitively expensive in training and inference.
It also complicates the deployment of such CNNs, especially on devices with limited memory and computing resources like autonomous vehicles, drones, and smartphones.

In recent years there has been an effort to improve the efficiency of CNNs.
Common approaches to reduce the number of weights in CNNs are pruning \cite{pruning92,SongHan2015,guo2016dynamic,PrunningCornel2017,Luo_2017_ICCV,chin2018layer}, sparsity \cite{wen2016learning,SparsConvCornell2017,SparseReguSongHan2016}, and quantization \cite{hubara2016binarized,li2017training,banner2018scalable}.
Pruning reduces the number of weights in the network after training.
The fact that in many cases large portions of the networks' weights can be removed with minimal reduction of its accuracy indicates a considerable redundancy and over-parameterization of standard CNNs~\cite{molchanov2016pruning}.
While pruning is effective in reducing the number of weights and floating point operations (FLOPs), it generally leads to an unstructured non-zero pattern of the weights, which increases the memory access costs.
The lack of structure also complicates the efficient deployment of the CNN on hardware.

Another approach to improve the efficiency of CNNs is to replace the fully-coupled convolution operators by sparser convolution operators (i.e., operators with fewer non-zero elements) before training.
One typical building block is known as a \emph{grouped} convolution operator, which partitions the channels into groups and only allow grouped coupling; see, e.g.,~\cite{KrizhevskySutskeverHinton2012}.
When the number of groups equals the number of channels, one obtains a \emph{depth-wise} convolution operator, which is a block diagonal matrix whose blocks are spatial convolution operators.
The depth-wise convolution operator filters each channel of the image data separately and thus restricts the interaction of each feature to its nearby features in the same channel.
It is common to use the depth-wise operator in conjunction with fully-connected point-wise $1\times1$ convolutions to introduce coupling across the channels.

A few CNN architectures have been derived using depth-wise and $1\times1$ convolution operators, often augmented with bottleneck or shuffling techniques; see, e.g.,~\cite{howard2017mobilenets,sandler2018mobilenetv2,BottleNeckMinWang,SuffleNet,ma2018shufflenet}.
These works use the depth-wise and $1\times 1$ convolution separately, with activation and batch normalization layers in between them.
Although in this paper we use $3\times3$ convolutional stencils to parameterize the depth-wise convolution, other choices are possible; in fact, mixing stencils of different sizes has shown promising results~\cite{Tan:2019vk}.
This typically requires a redesign of existing CNN architectures.
To reduce the ratio between FLOPs and memory access in depth-wise convolution operators, replacing convolutions with shifts has been proposed in~\cite{wu2018shift}.
Exploiting the multiscale structure of image data provides an alternative way to derive more efficient architectures; see the contemporary work~\cite{Chen:2019tp}.
It is known, however, that the memory access is often the real bottleneck in modern parallel hardware, and \emph{not necessarily FLOPs}. In fact, state-of-the-art implementations of depth-wise convolution operators on GPUs involve more FLOPs than necessary to achieve lower runtimes; see, e.g.,~\cite{qin2018diagonalwise}. Nevertheless, whether the dominant cost is the storage of the parameters, the FLOPs, or the memory access, is highly dependent on the hardware at hand. Therefore, it is desirable to build  convolution operators and architectures that are flexible in their definition so that they can be configured as necessary on any specific hardware.

\begin{figure}
    \centering
        \includegraphics[width=0.4\textwidth]{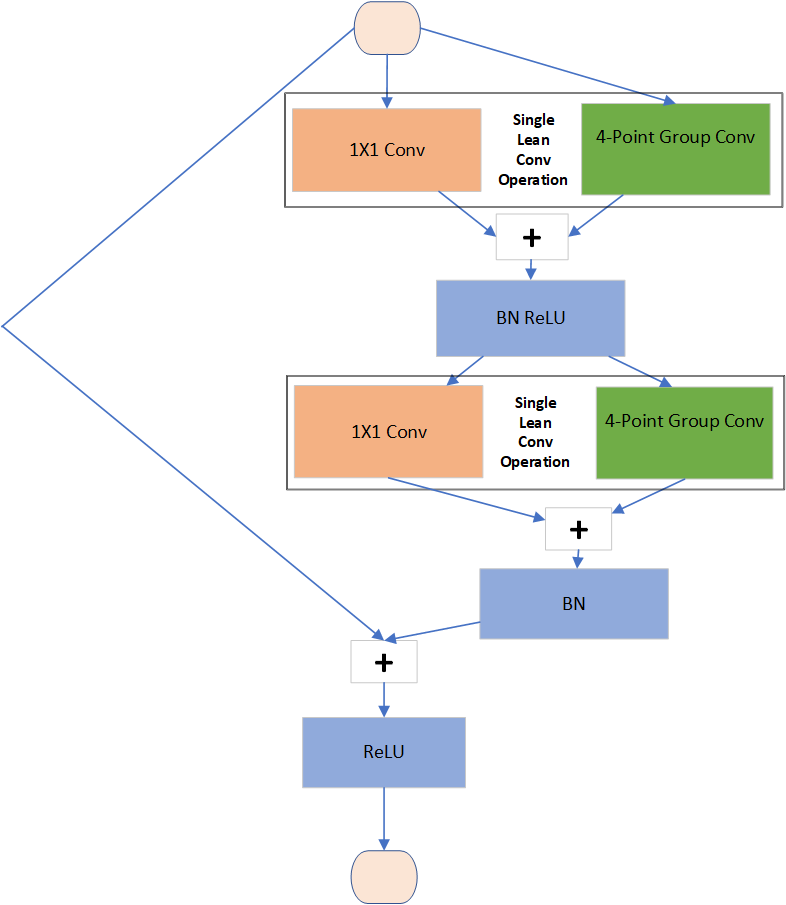}
    \caption{Building blocks of basic LeanResNet step. The $1\times1$ and spatial grouped convolutions are applied simultaneously.}
    \label{fig:LeanNetBlock}
\end{figure}

In this paper, we introduce LeanConvNets, a new family of CNNs built as lean versions of known networks, using lean convolution operators. These operators reduce the number of weights, computation time, and FLOPs while achieving competitive results. The lean operators preserve the overall network structure and can thus be applied to a variety of networks, e.g., residual networks (ResNets, ResNeXt) \cite{he2016deep,he2016identity,xie2017aggregated}, which have been two of the most reliable architectures in the literature. The following aspects set our work apart from other approaches: \newline $\bullet$ We obtain a new operator as the sum of the grouped and $1\times 1$ convolution operators (a schematic description of a ResNet block with lean convolutions appears in Fig. \ref{fig:LeanNetBlock}; more details later).  Using a prototype implementation, we show that handling both operations simultaneously reduces the computation time required to apply the operator. Also, this design introduces several opportunities for optimization in hardware through its parallelism, minimal number of memory accesses, and slightly reduced number of weights. Using grouped instead of depth-wise convolution operators allows one to gradually enlarge the portion of spatial convolutions in order to improve the performance of the lean networks. Our networks are mostly suitable to parallel devices that are bandwidth bounded and not computation bounded.  \newline  $\bullet$ We present two ways to reduce the spatial kernel size that further decrease the number of weights and FLOPs and are easy to implement efficiently. In the first method, we replace the standard $3\times3$ by a 5-point stencil. In the second method, we filter two-dimensional images using a one-dimensional convolution operator ($3\times1$ or $1\times3$, depending on memory layout) and its transpose applied at the memory write. This operator can be implemented with the same number of memory accesses as the $1\times1$ convolution since the memory of the feature maps is sequential in the one dimension. 

The remainder of the paper is organized as follows:
In Sec.~\ref{sec:prelim}, we discuss existing convolution operators and their computational costs in the context of residual neural networks.
In Sec.~\ref{sec:lean}, we introduce a family of lean convolution operators, analyze their costs, and outline their implementation.
In Sec.~\ref{sec:experiments}, we provide extensive numerical evidence for the efficacy of the resulting LeanConvNets for image classification and semantic segmentation.
In Sec.~\ref{sec:discussion}, we summarize the paper and discuss directions for future research.

\section{Preliminaries and notation}
\label{sec:prelim}
We now introduce our main notation and define the supervised classification and semantic segmentation problems that we use to validate our methods; for more details see~\cite{Goodfellow:2016wc}.
For brevity, we restrict the discussion to images although the techniques derived here can also be used for other structured data types such as audio or video data.
In supervised learning, we are given a set of training data consisting of pairs, $\{(\bfy_0^{(k)}, \bfc^{(k)})\}_{k=1}^s \subset \R^{\nf}\times \R^{\nc}$.
In our case, $\bfy_0^{(k)}$ is the $k$-th input image and $\bfc^{(k)}$ either represents the probabilities for the entire image (in classification) or each pixel (in segmentation) to belong to one of the pre-defined classes.
Our goal is to define a neural network architecture and train its weights $\bftheta \in \R^{\nTheta}$ and the weights of a linear classifier, denoted by $\bfW \in\R^{\nc \times \nOut}$ and $\bfmu \in\R^{\nc}$, such that
\begin{equation*}
	\bfc^{(k)} \approx S(\bfW \bfy^{(k)}(\bftheta) +  \bfmu), \quad \text{ for all } \quad k=1,2,\ldots,s.
\end{equation*}
Here, $S$ is a softmax hypothesis function and $\bfy^{(k)}(\bftheta) \in\R^{\nOut}$ denotes the output of the network applied to the $k$th sample.

The learning problem can be phrased as a minimization problem of a regularized empirical loss function
\begin{equation*}
	\min_{\bftheta,\bfW,\bfmu} \frac1s \sum_{k=1}^s L(S(\bfW \bfy^{(k)}(\bftheta) + \bfmu), \bfc^{(k)}) + R(\bftheta,\bfW,\mu),
\end{equation*}
where $L$ is the cross entropy loss and $R$ is a regularization function.
The optimization problem is usually solved using variants of stochastic gradient descent (SGD); see the original work~\cite{RobbinsMonro1951} and the survey \cite{bottou2016optimization}.

As a baseline architecture, we consider residual networks (ResNet) \cite{he2016deep,he2016identity}, which have been successful in many imaging tasks.
Given a data sample, $\bfy_0$, the forward propagation through an $N$-layer ResNet is defined as
\begin{equation}\label{eq:Resnet}
    \bfy_{l+1} = \bfy_{l} + \bfF(\bftheta_{l},\bfy_l), \quad \text{for} \quad l=0,\ldots,N-1,
\end{equation}
where $\bftheta_{l}$ is the set of weights associated with the $l$-th layer and we define $\bfy(\bftheta) = \bfy_0$.
There are different choices for the nonlinear term in \eqref{eq:Resnet}, e.g.,
\begin{equation}\label{eq:F}
\bfF(\bftheta_l,\bfy_l) = \bfK_2(\bftheta_{l,2}) \sigma( {\cal N}(\bfK_1(\bftheta_{l,1}) \sigma({\cal N}(\bfy_l)))).
\end{equation}
Here, $\sigma(x) = \max\{x,0\}$ denotes an element-wise rectified linear unit (ReLU) activation function and the weights are partitioned into $\bftheta_{l,1}$ and $\bftheta_{l,2}$ that parameterize the two linear operators $\bfK_1$ and $\bfK_2$, respectively.
For brevity, we omit the weights of the normalization layer $\mathcal{N}$.

\begin{figure*}
    \centering
    \includegraphics[width=\textwidth]{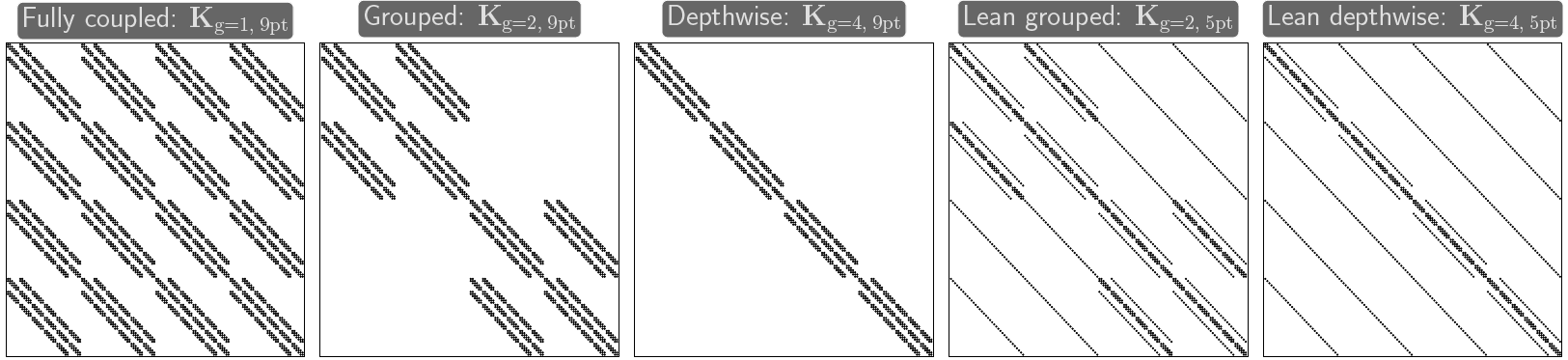}
    \caption{Sparsity patterns of different convolution operators for $6\times 6$ images with four input and output channels. The leftmost subplot shows the sparsity pattern of a $3\times3$ fully-coupled convolution operator. The next two subplots depict the grouped convolution operators for $g=2$ and $g=4$, respectively. The remaining two subplots show the proposed lean grouped and depth-wise operators that are a sum of a fully coupled $1\times1$ and a grouped (or depth-wise) spatial convolution operators.}
    \label{fig:spyplots}
\end{figure*}

In convolutional ResNets, the operators $\bfK_i$ in \eqref{eq:F} are composed of spatial convolution operators.
If the input $\bfy$ has $\cin$ channels, and the output $\bfK\bfy$ has $\cout$ channels, then the common choice for an operator $\bfK$	 is a $\cout \times \cin$ block matrix of convolutions, introducing full coupling across the channels.
For example, if $\cin=\cout=4$, then the convolution operators in \eqref{eq:F} can be written in matrix form as
\begin{equation}\label{eq:CNNfc}
	\bfK_{\rm full}(\bftheta)=\left(
		\begin{array}{@{}c@{\;\;}c@{\;\;}c@{\;\;}c@{}}
			\bfC^{(1,1)} &  \bfC^{(1,2)} &  \bfC^{(1,3)} &  \bfC^{(1,4)}\\
			\bfC^{(2,1)} &  \bfC^{(2,2)} &  \bfC^{(2,3)} &  \bfC^{(2,4)}\\
			\bfC^{(3,1)} &  \bfC^{(3,2)} &  \bfC^{(3,3)} &  \bfC^{(3,4)}\\
			\bfC^{(4,1)} &  \bfC^{(4,2)} &  \bfC^{(4,3)} &  \bfC^{(4,4)}
		\end{array}
	\right),
\end{equation}
where $\bfC^{(i,j)} = \bfC(\bftheta^{(i,j)})$ denotes the sparse matrix associated with the spatial convolution kernel parameterized by the $3\times3$ filter $\bftheta^{(i,j)}\in\mathbb{R}^{9}$.
For ease of notation, we do not explicitly denote the dependency on $\bftheta$ in the following.
The sparsity pattern of this operator is visualized in the leftmost subplot of Fig.~\ref{fig:spyplots} for an image size of $6\times6$. Applying $\bfK_{\rm full}$ requires $\mathcal{O}( \cin \cdot \cout)$ FLOPs, and $\bfK_{\rm full}$ has $9 \cdot \cin \cdot \cout$ weights.
In practice, each $\bfK_{\rm full}$ can have millions of weights.

\emph{Grouped convolutions} are popular alternatives to $\bfK_{\rm full}$ as they reduce the number of weights and computations.
In our example, we can restrict the interaction of the channels to $g=2$ groups, which leads to the block diagonal matrix
\begin{equation}\label{eq:CNNfcG}
	\bfK_{\rm g=2}=\left(
		\begin{array}{@{}cccc@{}}
			\bfC^{(1,1)} &  \bfC^{(1,2)} &  0& 0\\
			\bfC^{(2,1)} &  \bfC^{(2,2)} & 0& 0\\
			 0 &  0 &  \bfC^{(3,3)} &  \bfC^{(3,4)}\\
			 0 &  0 &  \bfC^{(4,3)} &  \bfC^{(4,4)}\\
		\end{array}
	\right).
\end{equation}
This reduces the number of weights and FLOPs by a factor of $\rm g$ compared to the full convolution.
Clearly, $\bfK_{\rm full} = \bfK_{g=1}$ and for $\rm g = \cin$, we get the depth-wise convolution; the sparsity pattern of  $\bfK_{\rm g=2}$ and $\bfK_{\rm g=4}$ are shown in Fig.~\ref{fig:spyplots}.
Grouped convolutions can be extended to rectangular operators when $g$ divides both $\cin$ and $\cout$.

\section{Lean Convolutional Operators}
\label{sec:lean}
We now introduce a family of lean convolutional operators that achieve competitive performance and reduce the number of weights, memory access, and FLOPs.
It has been shown that $1\times 1$ convolutions can be effective if complemented with a relatively small number of spatial convolutions~\cite{howard2017mobilenets,sandler2018mobilenetv2,BottleNeckMinWang,SuffleNet,ma2018shufflenet}.
Then, the computational cost of the $1\times 1$ convolution (in terms of FLOPs and weights) dominates the cost of the  spatial convolutions as the number of channels grows.
It has also been observed that the accuracy of the network suffers from the relative shortage of spatial convolutions, which is often explained by a relatively small number of weights.
To increase the accuracy, our lean convolution operators aim to allocate the weights more efficiently between grouped and $1\times1$ convolutions.
To this end, we reduce the kernel size on the one hand and add spatial convolutions on the other.
The group size of the spatial convolution is a hyper parameter that trades off between accuracy and computational efficiency. This also allows us to accommodate different computational devices without changing the high-level structure of the network.

We obtain lean convolution operators in three steps.
First, lean operators are a sum of $1\times1$ and grouped spatial convolutions---see Fig. \ref{fig:LeanNetBlock}. If implemented efficiently the lean convolution allows one to reuse of memory access, increase parallelism, and further reduce the number of weights.
Second, the group size parameter $g$ allows the user to balance between spatial filtering (by a grouped operator) and coupling (by $1\times1$ operators) and control the performance of the network.
Third, we use convolutional filters with only five or three elements instead of 9 for common $3\times3$ filters, which reduces the number of weights and FLOPs.

Continuing our example from above,  we define the lean analogue to~\eqref{eq:CNNfc}-\eqref{eq:CNNfcG} as
\begin{equation}\label{eq:Klean_g2}
	\bfK_{\rm lean, g=2} = 	\left(
		\begin{array}{@{}cccc@{}}
			 \bfC^{(1,1)}     & \bfC^{(1,2)}      & \alpha_{1,3}\bfI & \alpha_{1,4}\bfI\\
			 \bfC^{(2,1)}     & \bfC^{(2,2)}      & \alpha_{2,3}\bfI & \alpha_{2,4}\bfI   \\
			 \alpha_{3,1}\bfI & \alpha_{3,2}\bfI  & \bfC^{(3,3)}     & \bfC^{(3,4)}\\
			 \alpha_{4,1}\bfI & \alpha_{4,2}\bfI  & \bfC^{(4,3)}     & \bfC^{(4,4)}\\	
		 \end{array}
	\right),
\end{equation}
where $\bfI$ is a scaled identity matrix and $\alpha_{i,j}\in\R$ are weights.
The identity operators represent the $1\times 1$ convolution and the convolution operators $\bfC^{(i,j)}$ enable spatial filtering. Similarly, the lean operator with $g=4$ groups is
\begin{equation}\label{eq:Klean}
	\bfK_{\rm lean, g=4} = 	\left(
		\begin{array}{@{}cccc@{}}
			\bfC^{(1,1)} & \alpha_{1,2}\bfI   &\alpha_{1,3}\bfI &\alpha_{1,4}\bfI \\
			 \alpha_{2,1}\bfI & \bfC^{(2,2)} &\alpha_{2,3}\bfI &\alpha_{2,4}\bfI   \\
			 \alpha_{3,1}\bfI & \alpha_{3,2}\bfI  &\bfC^{(3,3)}& \alpha_{3,4}\bfI\\
			 \alpha_{4,1}\bfI & \alpha_{4,2}\bfI  &\alpha_{4,3}\bfI &\bfC^{(4,4)}\\	
		 \end{array}
	\right),
\end{equation}
The sparsity patterns of these operators are shown in fourth and fifth subplots of Fig.~\ref{fig:spyplots}.

The setting in \eqref{eq:Klean} with $\rm g =4$ can be seen as the sum  of depth-wise and $1\times1$ convolutions, which are also used in \cite{howard2017mobilenets, sandler2018mobilenetv2,ma2018shufflenet}.
These works perform the depth-wise and $1\times1$ convolutions separately---the depth-wise convolution is applied between two $1\times 1$ convolutions with ReLU operations between them.
Since we sum the operators we can apply them both simultaneously, which allows us to optimize memory access and improve parallelism (more can be done at once).

\subsection{The argument for groups in compact networks}
Our experiments suggest that if we take a given compact network that utilizes depth-wise and $1\times1$ operations, and define its \emph{full} version by placing a $3\times3$ convolution instead of each $1\times1$ convolution, we get networks that are significantly more expensive, but perform better in terms of accuracy.
Employing such compact schemes may result in a spatial component that is too small, especially when the number of channels is large and the $1\times1$ operators dominate the spatial convolutions.
This motivates us to add a small number (e.g., $\frac{\cin\cout}{\rm g}$) of spatial convolutions to improve the performance of the network compared to depth-wise operators.

The motivation for using this operator is as follows: first, the implementation of the grouped convolution works best in groups of intermediate size, and it is often even more efficient to zero-pad the groups (artificially enlarge them) to get better computational performance on GPUs~\cite{qin2018diagonalwise}. Second, in the standard combination of depth-wise and $1\times1$ convolutions, the former becomes negligible compared to the the latter as the number of channels grow, hurting accuracy without providing considerable savings.
Our proposal in this context is to use the grouping mechanism to keep a constant ratio of operations between the two types of convolutions, such that the $1\times1$ convolution that has $\cin\cdot\cout$ weights is more dominant than the grouped convolution that has $\frac{(r-1)\cin\cout}{\rm g}$ weights\footnote{The cost in operations is proportional to the number of weights.}, where $r$ is the stencil size (e.g., $r=9$ for a $3\times3$ stencil).
For example, if we choose a ratio of $\frac{1}{8}$, then we set $\rm g \approx 8(r-1)$
s.t. the number of channels is divisible by $\rm g$.
We subtract 1 from $r$ since the middle weight is included in the $1\times1$ convolution.

We note that enhancing the lean convolution with the grouping mechanism is similar to enhancing the depth-wise convolution in networks such as MobileNetV2.
This would result in a network that is similar to the ResNeXt networks \cite{xie2017aggregated}, which applies a grouped convolution instead of the full $3\times3$ convolution in the bottleneck version of the original ResNet.
The grouping helps to enlarge the bottleneck expansion while keeping a low additional cost, and without adding many weights.
Although the works were proposed independently, maximizing the number of groups in ResNeXt leads to a network which is similar to MobileNetV2.

\subsection{LeanConv 5-pt: lean convolutions with 5-point stencils.}

The first version of LeanConvNets is based on 5-point convolution stencils.
The idea is to replace the stencils of $\bfC^{(i,j)}$ in~\eqref{eq:Klean_g2} and~\eqref{eq:Klean} by the 5-point stencil
\begin{equation}\label{eq:5point}
 \left[\begin{matrix}
0 & c_{i,1} & 0 \\ c_{i,2} & \alpha_{i,i} &   c_{i,3} \\
0&  c_{i,4} &0
\end{matrix}\right],
\end{equation}
where $\alpha_{i,i}$ is the $i,i$-th entry of the $1\times1$ convolution, and $c_{i,1},...,c_{i,4}$ are additional four weights per input channel $i$.
An example for the sparsity pattern of the resulting lean operator with 5-point convolutions is shown in Fig.~\ref{fig:spyplots}. The operator $\bfK_{{\rm lean},\rm g}$ with 5-point stencils and $\rm g$ groups has $(1+ \frac{4}{\rm g})  (\cin\cdot \cout)$ weights. We note that if the number of channels and $\rm g$ are large, then the $1\times1$ convolution is the dominating operator both in terms of weights and FLOPs.

The lean convolution can replace fully-coupled convolution operators in many existing CNNs without any structural changes to the architecture.
A straightforward way to implement a grouped lean convolution like in \eqref{eq:Klean_g2} is to use the package {\tt cudnn}  to perform the $1\times1$ and spatial convolutions separately.
As we show later, our custom implementation, which simultaneously applies both operations, outperforms the {\tt cudnn} approach for $\rm g = c_{in}$.

To motivate the use of 5-point stencils, consider the ResNet architectures that have been recently interpreted as time-dependent nonlinear ordinary differential equations (ODEs); see, e.g., \cite{haber2017stable,Chang2017Reversible,E2017,ChaudhariEtAl2017,lu2018beyond,chen2018neural,haber2019imexnet}.
This allows the community to analyze and extend ResNets using theoretical and practical ideas from the world of ODEs and PDEs~\cite{ruthotto2018deep}.
In this point of view, the five-point stencil is able to express a mass term, and a discretization of first and second spatial derivatives in the $x$ and $y$ dimensions. That is, the first and second derivatives in the $x$ dimension can be approximated by
\begin{equation}\label{eq:derivatives}
\frac{\partial}{\partial x} \approx \frac{1}{2h_x}[-1, 0, 1]\quad\text{and}\quad \frac{\partial^2}{\partial x^2} \approx \frac{1}{h_x^2} [1, -2, 1],
\end{equation}
where $h_x$ is the edge length of a pixel.
This, together with $\frac{\partial}{\partial y}$, $\frac{\partial^2}{\partial y^2}$ and the mass term (or 5-point low-pass filter) are included in the span of the 5-point stencil \eqref{eq:5point}.
The remaining four entries of a full $3\times3$ stencil correspond to mixed partial derivatives, which rarely occur in PDE-models, and are thus good candidates for reducing computations and weights.

\bigskip
\emph{Implementation}: The standard $3\times3$ convolution is implemented using a shift per stencil parameter (known as the {\tt shiftIm2col} operation), and a matrix-matrix multiplication using the function {\tt gemm}.
In the same way we can multiply a 5-point convolution operator, trivially saving $4/9$ of the operations.
For small group sizes, and in particular for the depth-wise setting (group size of 1), we found that a direct implementation of the convolutions is faster than the standard implementation with shiftIm2col. As the groups get larger, the approach using shiftIm2col is more preferable, due to the efficiency of {\tt gemm}. Even more savings can be realized in 3D CNNs where the standard 27-point stencils are replaced with 7-point stencils (the 3D version of \eqref{eq:5point}), saving $20/27$ of the operations and weights.

\subsection{LeanConv 3-pt: lean convolutions with 1D 3-point stencils}

In this section, we present a more sophisticated lean convolution operator that can be applied almost at the same cost as a $1\times1$ convolution as both operators use the same memory accesses.
This convolution is based on $1D$ convolution operators, either $1\times3$ or $3\times1$, which can be applied efficiently if the memory is continuous in the direction of the 1D kernel.

In addition to the benefits from an implementation perspective, the use of 1D kernels can also be motivated as follows:
It is known that in 2D, a large portion of the $3\times3$ kernel can be parameterized by a multiplication of  $1\times3$ and $3\times1$ kernels, also called \emph{separable} kernels.
Separable kernels can represent many of the important operators, such as low-pass filters, and the spatial derivatives in \eqref{eq:derivatives}.
Our idea is to use two convolutions, such as $\bfK_1$ and $\bfK_2$ in \eqref{eq:F}, to effectively apply separable operators: $\bfK_1$ applies a $1\times 3$ kernel in the horizontal direction, and $\bfK_2$ applies a $3\times 1$ kernel in the vertical direction.
We note that 1D stencils were also used in a small section of the InceptionV4 network \cite{szegedy2017inception}.
There, $1\times7$ and $7\times1$ were used in together with $3\times3$ convolutions to increase the field of view of the network.
Here we show that even if we use  $1\times3$ and $3\times1$ convolutions \emph{only}, we can still get an effective network, while reducing the number of weights, FLOPs and (most importantly) memory access. The latter can be saved if the memory of the feature maps is aligned with the direction of the kernel.
Our idea here to maintain the memory alignment is to apply the convolutions together with channel transposition. That is, if the 1D convolution operator $\bfK_1$ is aligned with the memory, then the feature maps are transposed during the WRITE operation to prepare the result to $\bfK_2$ that is aligned in the other direction, and vice-versa.

\bigskip
\emph{Custom GPU Implementation:}
To explain the 3-pt lean convolution, we first briefly describe one of the approaches for multiplying matrices on a GPU---that is essentially the $1\times1$ convolution operator.
We follow the description of the {\tt cutlass} library \cite{cutlas}, and the implementation of the MAGMA open source project \cite{magma}.
 Given two matrices $\bfK\in\mathbb{R}^{\cout\times\cin}$ and $\bfY\in\mathbb{R}^{\cin\times n}$, we first divide their product $\bfK\bfY\in \mathbb{R}^{\cout\times n}$ into tiles of size $t_n\times t_o$.
 Each of these tiles is computed by a multiplication of a block of $t_o$ columns of $\bfK$ and $t_n$ columns of $\bfY$.
These sub-matrices are also divided into sub-blocks of size $t_i$. Each group of physical cores gets a task of computing a tile of $t_n\times t_o$ output numbers. To apply this, we first fetch the relevant tiles into shared memory, and then multiply them in parallel. Algorithm \ref{alg:TiledMuliplication2} summarizes the procedure, ignoring the underlined parts; see \cite{cutlas,magma} for more details.

Now we explain how to apply the 3-pt lean operator, assuming that the number of convolutions is small compared to $\cin\times \cout$. An important consideration for GPU implementations is that fetching memory from global memory into shared memory is slow, while accessing the shared memory is fast.
Our idea is to add a small memory fetch into the procedure above, and apply the 1D convolution to the already-fetched tile of $\bfY$, assuming that the memory of $\bfY$ is continuous in the same direction of the 1D kernel.

To finish the operation, we now wish that the direction of the next kernel is aligned with the direction of the data. This will be true only for one direction, and we handle that by transposing the feature maps during the \emph{write} phase at the end of the convolution.
Thus, when multiplying $\bfK_1$ in \eqref{eq:F} we have the maps aligned in one direction, but during the multiplication, we transpose the data in shared memory and write it transposed. After a ReLU operation (for which the direction does not matter) the input to $\bfK_2$ is ready to be multiplied and is aligned in the other direction. At the end of the same multiplication of $\bfK_2$, the result is again transposed back to the original alignment.
Algorithm \ref{alg:TiledMuliplication2} summarizes the 3-pt lean convolution procedure. Compared to the 5-pt lean convolution, the 3-pt conv requires that at least two kernels are applied one after the other before a skip connection, such that the maps are transposed back to their original form. We note that this algorithm is mostly beneficial with large number of groups, and in particular with depth-wise configuration. If the groups are large, the shiftIm2Col approach is preferable.

\begin{algorithm}
\caption{Tiled 3-point LeanConv Multiplication}\label{alg:TiledMuliplication2}
\begin{algorithmic}[1]
\State \emph{\# Computation of $\bfX=\bfK\bfY$.}
\State \emph{\# $t_n$,$t_o$,$t_i$: tile sizes. $(i,j)$: thread id.}
\State \emph{\# The underline parts add over simple $1\times1$ conv}.
\Procedure{Lean3ptGEMM}{$\bfX,\bfK,\bfY,i,j$}
\State \underline{Fetch boundary values from $\bfY$ and spatial}
\State $\;\;\;$ \underline{convolution parameters into shared memory $C$.}
\For{$k=1,...,\lceil\frac{\cin}{t}\rceil$}
\State \emph{\# Each thread in the tile fetches two blocks:}
\State Fetch tile $i,k$ from $\bfK$ to shared $A\in\mathbb{R}^{t_n\times t_i}$.
\State Fetch tile $k,j$ from $\bfY$ to shared $B\in\mathbb{R}^{t_i\times t_o}$.
\State \emph{\# Multiply $AB$ by $t_i$ outer products:}
\State Multiply $AB$ into local memory.
\State \underline{If relevant to the output tile:}
\State $\;\;\;$ \underline{Apply convolution to $B$ into local memory.}
\State Write \underline{transposed} local memory to $\bfX$.
\EndFor
\EndProcedure
\end{algorithmic}
\end{algorithm}

\section{Experiments} 
\label{sec:experiments}

\begin{table}
\centering
\small
\begin{tabular}{|c|c|c|c|}
\hline
Type & Layer width (channels) & $\#$ Steps & Strides\\
\hline
Res18 &32-64-128-256 & 2-2-2-2 & 1-2-2-2\\
Res24-narrow & 12-24-48-96 & 2-3-3-3 & 1-2-2-2\\
Res24 & 32-64-128-256 & 2-3-3-3 & 1-2-2-2\\
Res34 &64-128-256-512 & 3-4-6-3 & 1-2-2-2\\
Res38-narrow &24-48-96-192-384 & 4-5-5-3-1 & 1-2-2-2-2\\
Res38 &64-128-256-512-1024 & 4-5-5-3-1 & 1-2-2-2-2\\
Res40-narrow & 24-48-96-192 & 3-5-7-4 & 1-2-2-2\\
Res40 & 64-128-256-512 & 3-5-7-4& 1-2-2-2\\

\hline
\end{tabular}
\caption{Network configurations.}
\label{tab:config}
\end{table}

\begin{table*}
    \centering
    \small
	\begin{tabular}{|l|ccc|ccc|}
\hline
                  & \multicolumn{3}{c|}{CIFAR10} & \multicolumn{3}{c|}{CIFAR100}\\
Architecture      & Network & Params$\setminus$FLOPs[M] & Test acc.  & Network &Params$\setminus$FLOPs[M] & Test acc.  \\
\hline
 ResNet          &  Res24  & 4.7$\setminus$212 &  94.5\%  & Res40 & 28.9$\setminus$1490 &   78.5\% \\
 ResNet (small)	 &  Res24-narrow    & 0.66$\setminus$53&   92.0\%  &Res40-narrow & 3.8$\setminus$239 &  72.3\%  \\
 MobileNetV2     &Res24$*$      & 0.50$\setminus$33 &  91.7\%  & Res40$_{*}$   & 3.1$\setminus$167&  71.9\%\\
 ShuffleNetV2    & 0.5x     &0.35$\setminus$42 &  91.6\%   & 1.5x  & 2.6$\setminus$375& 74.2\%\\
 ShiftResNet        &Res24$*$      & 0.49$\setminus$31 & 90.7\%  &Res40$_{*}$& 3.1$\setminus$201 &   74.2\%\\
 LeanResNet  5-pt$_{DW}$ [ours]      &  Res24   &   0.53$\setminus$26& 92.8\%   &Res40&3.3$\setminus$167&  74.3\%  \\
  LeanResNet 5-pt$_{g=16}$ [ours]      & Res24&0.65$\setminus$31 &  93.7\% &Res40& $4.0\setminus$203&75.7\% \\
 LeanResNet 3-pt$_{g=8}$ [ours]      & Res24& 0.66$\setminus$31 &93.4\%  &Res40&  4.1$\setminus$203&75.3\% \\
 \hline
\hline
                  & \multicolumn{3}{c|}{STL10} & \multicolumn{3}{c|}{TinyImageNet200}\\
Architecture      & Network & Params$\setminus$FLOPs[M] & Test acc.  & Network &Params$\setminus$FLOPs[M] & Val. acc.  \\
\hline
 ResNet          & Res24& 4.7$\setminus$1908 &  86.6\% & Res38&40.9$\setminus$4816&65.2\%\\
 ResNet (small)	  &Res24-narrow& 0.66$\setminus$277& 82.5\% &Res38-narrow & 5.8$\setminus$831&61.3\%\\
 MobileNetV2     &  Res24$_{*}$ &0.50$\setminus$302 & 84.0\%\ &1.4 \cite{sandler2018mobilenetv2}& 4.7$\setminus$661& 56.4\%\\
 ShuffleNetV2   & 1.0x  &1.2$\setminus$608 & 81.7\% & 2.0X \cite{ma2018shufflenet} & 5.7$\setminus$740 & 58.4\%\\
 ShiftResNet       & Res24$_{*}$  & 0.49$\setminus$361 & 84.0\% & Res38$_{*}$ & 4.5$\setminus$793 & 61.8\%\\
 LeanResNet 5-pt$_{DW}$ [ours]      & Res24& 0.53$\setminus$235&84.0\% &Res38& 4.7$\setminus$488 & 62.6\%\\
 LeanResNet 5-pt$_{g=16}$ [ours]      & Res24&0.65$\setminus$275&  86.5\% &Res38& 5.9$\setminus$590&63.4\% \\
 LeanResNet 3-pt$_{g=8}$ [ours]      & Res24& 0.66$\setminus$275 &85.4\%  &Res38&  5.9$\setminus$590&63.4\% \\

 \hline
\end{tabular}
	\caption{Comparison of classification results for small datasets. To make a fair comparison, we seek to match the number of parameters and FLOPs for each network. For MobileNetV2 and ShiftResNet we use expansion of $\epsilon=6$, and choose the width (number of channels) to be approximately $\sqrt{6}$ smaller than the width of LeanResNet, so that their number of parameters and FLOPs are comparable. We denote this by $*$.
	Since the images here are smaller that those of ImageNet, we removed the strides from the first steps of the networks MobileNetV2 and ShuffleNetV2.}
	\label{tab:Classification}
\end{table*}

\begin{table*}
    \centering
    \small
	\begin{tabular}{|l|ccc|}
\hline
                  & \multicolumn{3}{c|}{ImageNet} \\ 
Architecture      & Network & Params$\setminus$FLOPs[M] & Val. acc.\\  
\hline
 Resnet & Res34 \cite{he2016deep} & 21.8$\setminus$3600 & 74.0\%\\
 ResNeXt & Res50 \cite{xie2017aggregated} & 25.0$\setminus$4100 & 77.8\%\\
 LeanResNeXt 5-pt$_{g=32}$          &  Res34  & 3.6$\setminus$680 &  71.7\%   \\
 LeanResNeXt 5-pt$_{g=16}$          &  Res34  & 3.9$\setminus$630 &  72.1\%   \\
 MobileNetV2     &   1.0 \cite{sandler2018mobilenetv2}& 3.47$\setminus$301& 71.9\%\\
 LeanMobileNetV2 5-pt$_{DW}$&   1.0 & 3.49$\setminus$310& 72.2\%\\
 ShuffleNetV2   & 1.5x \cite{ma2018shufflenet} & 3.5$\setminus$299 & 72.6\%\\
 ShiftResNet  & ShiftNet-A\cite{wu2018shift}  & 4.1$\setminus$- & 70.1\%\\
 \hline
\end{tabular}
	\caption{Comparison of classification results for ImageNet using different compact networks. 
	}
	\label{tab:ClassificationImageNet}
\end{table*}

We demonstrate the proposed LeanConvNet approach and compare the lean versions of ResNet and ResNeXt \cite{xie2017aggregated}, called ``LeanResNet'', and ''LeanResNeXt'' to a fully-coupled ResNet, and other recent state-of-the-art compact architectures: ShuffleNetV2 \cite{ma2018shufflenet}, MobileNetV2 \cite{sandler2018mobilenetv2}, and ShiftResNet \cite{wu2018shift}.
We consider the image classification and semantic segmentation tasks using several data sets.
Our primary focus is to compare how different architectures perform using a relatively small number of parameters and FLOPs (we count floating point multiplications). Our experiments are performed with the PyTorch software \cite{paszke2017automatic}. In a third experiment, we also show that our lean operators can be implemented efficiently and, for $g=\cin$, outperform the separate application of depth-wise and $1\times1$ operators using the highly optimized package {\tt cudnn}.

As our focus is on the performance of the lean convolution operators, we use the established ResNet architectures as baseline for comparison, and we use the same structure of those ResNets, only with lean convolutions.
Our ResNet networks consist of several blocks that are preceded by an opening convolutional $3\times3$ layer, which initially increases the number of channels.
Then, there are several blocks, each consisting of a ResNet-based steps like Eq. \eqref{eq:Resnet}.
Each convolution is followed by ReLU and batch normalization as described in \eqref{eq:Resnet}. To increase the number of channels and to down sample the image, we concatenate the feature maps with a depth-wise convolution applied to the same channels, thus doubling the number of channels. This is followed by an average pooling layer. The last block consists of a pooling layer that averages each channel's feature map to a single pixel, and we use a linear classifier with softmax and cross entropy loss.

Although the architectures of LeanResNets and ResNet are similar, the former employs efficient convolutions such as \eqref{eq:Klean}.
The convolution sizes of MobileNetV2 and ShiftResNet were chosen such that the size of each expanded (by 6) $1\times 1$ convolution is equivalent to the size of a square $1\times1$ convolution of LeanResNet.
The architecture ShuffleNetV2 is evaluated with the configurations (0.5x,1.0x,1.5x,2.0x) in~\cite{ma2018shufflenet}.

For the classification of the ImageNet dataset (Table \ref{tab:ClassificationImageNet}), we used the ResNeXt configuration \cite{xie2017aggregated}, which starts with an opening layer of a strided $7\times7$ convolution followed by max pooling. Its basic step is combined from 3 convolutions (with ReLU and normalization), where the middle is  a grouped $3\times3$ and the rest are $1\times1$ convolutions. To limit the number of parameters in our LeanResNeXt, we replace each $1\times1$ convolution with the lean convolution, and the $3\times3$ convolution with a 5-pt convolution. The last block is identical to that of ResNet. In this ResNeXt architecture, we  reduced the number of channels compared to the original network, since its cost is dominated by $1\times1$ convolutions. Also, we use a lean version of MobileNetV2, where all the sizes of the network are identical to the original network, except we use 5-pt lean convolutions with maximal number of groups instead of the $1\times1$ convolutions in the basic step. This hardly adds parameters or FLOPs to the network.

\subsection{Image Classification}

We consider the CIFAR10, CIFAR100, STL10, ImageNet, and tinyImageNet200 datasets.
The CIFAR-10/100  datasets~\cite{krizhevsky2009learning} consists of 60k natural images of size $32\times32$ with labels assigning each image into one of ten categories (for CIFAR10) or 100 categories (for CIFAR100). The data are  split into 50K training and 10K test sets. The STL-10 dataset~\cite{coates2011analysis} contains 13K color-images each of size $96\times96$ that are divided into 5K training and 8K test images that are split into the ten categories. The  ImageNet \cite{ImageNet} challenge ILSVRC consists of over 1.28M images of size $224\times 224$ with labels assigning each image into one of 1000 classes where each class has 50 validation images.
The tinyImageNet200 \cite{tiny200} is a subset of the ImageNet dataset, and consists of 110K labeled images of size $64\times 64$ belonging to 200 classes, where each class has 500 training images and 50 validation images.  

For each of the data sets we used a different configuration, according to the difficulty of that data set. Table \ref{tab:config} summarizes the network weights that we use, which differ in the number of channels and the number of repetitions for each layer. All the networks are trained from scratch, i.e., no pre-trained weights are used. 
\begin{figure}
\vspace{-15pt}
    \centering
        \includegraphics[width=0.48\textwidth]{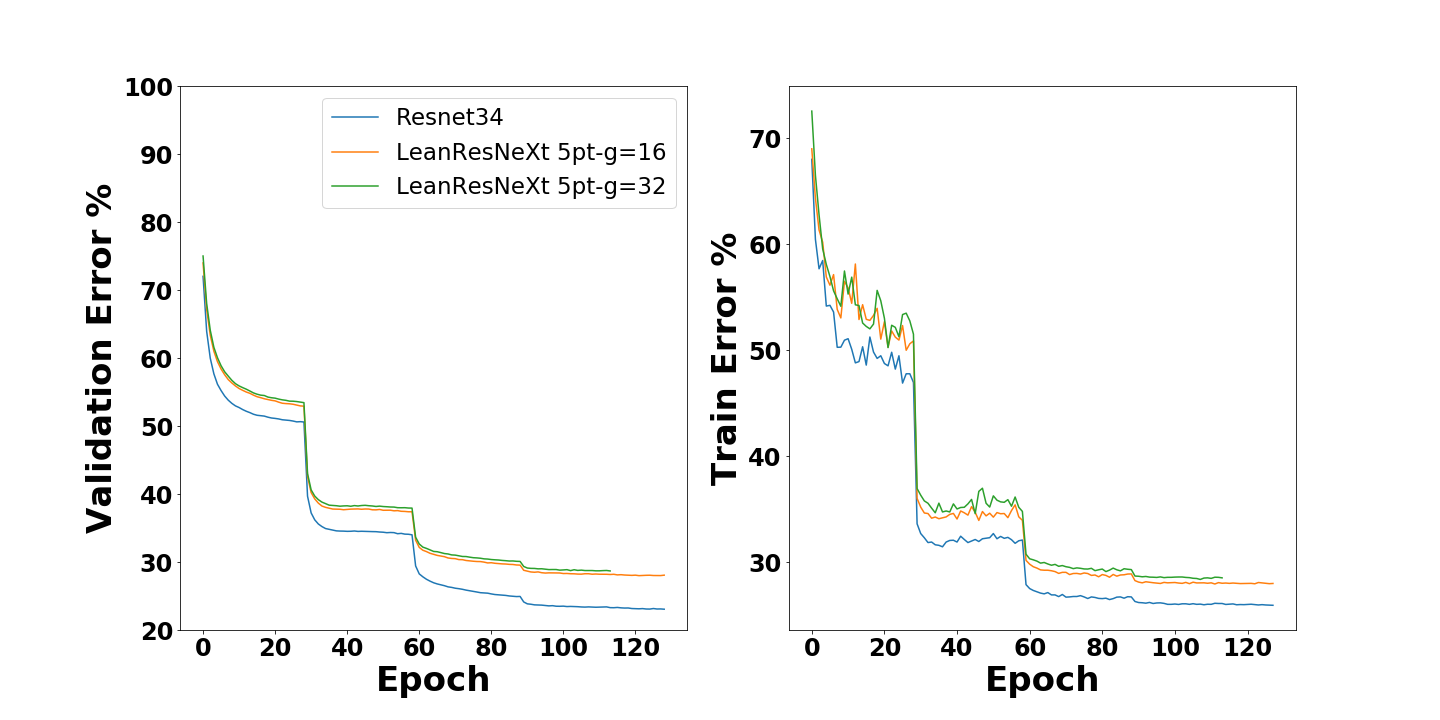}
    \caption{Validation (left) and Train (right) error per epoch for the ImageNet dataset.}
    \label{fig:ConvergencePlot}
\end{figure}
As optimization strategy for TinyImageNet200 we
use momentum SGD with a mini-batch size of $64$ for $300$ epochs. The learning rate start at $0.05$ and is reduced to $0.01$, $0.005$ and $0.001$ after the epochs $75$, $150$ and $225$ respectively. The weight decay is $0.0001$ and the momentum is $0.9$. The strategy for the other data sets is similar, with slight changes in the number of epochs, batch sizes and the timing for reducing the learning rate. We use standard data augmentation, i.e., random resizing, cropping and horizontal flipping.

Our classification results are given in Tables \ref{tab:Classification}-\ref{tab:ClassificationImageNet}, where we chose several representative configurations of groups for the lean convolutions.  The results show that our architecture is on par with and in some cases better than other networks. There is no preferred architecture between all options, but our architecture has the advantage of simplicity and resemblance to a standard and reliable ResNet network, which, as expected, yields better accuracy than all the other network at the expense of more parameters and cost. In Fig. \ref{fig:ConvergencePlot} we show the training and validation convergence plots of the architectures for the ImageNet. The plots show that the convergence of the LeanResNeXt is similar to that of the standard ResNet. 


\begin{table*}
    \centering
    \small
	\begin{tabular}{|l|c|ccc|ccc|}
\hline
                &  & \multicolumn{3}{c|}{CIFAR10} & \multicolumn{3}{c|}{CIFAR100}\\
Architecture      & Groups & Network & Params $\setminus$ FLOPs[M]  & Test acc.  & Network &Params $\setminus$ FLOPs [M]& Test acc.   \\
\hline
 ResNet 9-pt        &     ---        &  Res18  & 2.7$\setminus$181 & 94.3\%  & Res34 & 21.1$\setminus$1325& 77.6\% \\
  ResNet 5-pt    &   --- & Res18 & 1.5$\setminus$101 &94.0\% &Res34 &11.8$\setminus$739 & 78.0\%  \\
  ResNet 3-pt    &   ---  & Res18 & 0.92$\setminus$62 & 93.4\% &Res34 &7.1$\setminus$445  & 76.5\%     \\
  \hline
 LeanResNet 9-pt & DW (g = $\cin$) &Res18 &0.33$\setminus$25 & 91.1\%&Res34 &2.5$\setminus$160 & 73.0\% \\
 LeanResNet 9-pt & g = 32 &Res18 & 0.39$\setminus$27 & 91.6\%&Res34 &3.0$\setminus$188 & 74.8\% \\
 LeanResNet 9-pt & g = 16 &Res18 & 0.46$\setminus$32 & 92.0\%&Res34 &3.6$\setminus$225 & 75.6\% \\
 LeanResNet 9-pt & g = 8 &Res18 & 0.62$\setminus$42 & 92.9\% &Res34 &4.8$\setminus$ 298 & 76.7\% \\
 LeanResNet 9-pt & g = $\cin/32$  &Res18 & 0.80$\setminus$107 & 93.6\% &Res34 &4.3$\setminus$426 & 76.5\% \\
 LeanResNet 9-pt & g = $\cin/16$  &Res18 & 0.56$\setminus$64 & 93.0\% &Res34 &3.4$\setminus$289 & 75.5\% \\
 LeanResNet 9-pt & g = $\cin/8$ &Res18 & 0.43$\setminus$43 & 92.5\% &Res34 &2.9$\setminus$220 & 74.9\% \\
 \hline
 LeanResNet 5-pt &DW (g = $\cin$) &Res18 & 0.32$\setminus$23 & 91.0\% &Res34 &2.4$\setminus$156 & 72.7\% \\
 LeanResNet 5-pt & g = 32 &Res18 & 0.35$\setminus$25 & 91.7\%&Res34 &2.7$\setminus$170 & 75.0\% \\
 LeanResNet 5-pt & g = 16 &Res18 & 0.39$\setminus$27 & 92.5\% &Res34 &3.0$\setminus$188 & 75.7\% \\
 LeanResNet 5-pt & g = 8  &Res18 & 0.46$\setminus$32 & 92.8\%&Res34 & 3.6$\setminus$225 & 76.1\% \\
LeanResNet 5-pt & g = $\cin/32$ &Res18 & 0.56$\setminus$64 & 93.9\% &Res34 & 3.4$\setminus$289 & 76.5\% \\
 LeanResNet 5-pt & g = $\cin/16$ &Res18 & 0.43$\setminus$43& 93.2\% &Res34 & 2.9$\setminus$220 & 75.7\% \\
 LeanResNet 5-pt & g = $\cin/8$ &Res18 &0.37$\setminus$33 & 92.8\% &Res34 & 2.6$\setminus$186 & 75.6\% \\
 \hline
 LeanResNet 3-pt &DW (g = $\cin$)   &Res18 & 0.31$\setminus$23 & 90.5\% &Res34 &2.4$\setminus$153 &72.7\%\\
 LeanResNet 3-pt & g = 32 &Res18 &0.33$\setminus$23 & 91.0\% &Res34 &2.6$\setminus$160 &73.9\%\\
 LeanResNet 3-pt & g = 16 &Res18 & 0.35$\setminus$24& 91.4\%&Res34 &2.7$\setminus$170 &74.5\%\\
 LeanResNet 3-pt & g = 8  &Res18 & 0.39$\setminus$27& 92.4\% &Res34 &3.0$\setminus$188 &74.8\%\\
 LeanResNet 3-pt &$\cin/32$  &Res18 &0.43$\setminus$43& 92.5\%  &Res34 &2.9$\setminus$220 & 76.5\%\\
 LeanResNet 3-pt & $\cin/16$  &Res18 & 0.37$\setminus$33 & 92.2\% &Res34 &2.6$\setminus$186 & 75.0\% \\
 LeanResNet 3-pt & $\cin/8$  &Res18 & 0.35$\setminus$27 & 92.0\%&Res34 &2.5$\setminus$169 & 74.3\%\\
\hline
	\end{tabular}
		\caption{Classification results for the CIFAR10/100 datasets. Keeping the same basic architectures we study the impact on groups and stencil sizes on the test accuracy.}
	\label{tab:ClassificationLeanGroups}
\end{table*}

\emph{The influence of groups and stencils size:}
In this set of experiments we demonstrate the classification accuracy of LeanResNet with different configuration of grouping and stencil sizes on CIFAR10 and CIFAR100 data sets. We use small networks so that the differences in performance are more obvious. Table \ref{tab:ClassificationLeanGroups} presents the classification results. The configuration of $\rm g = \cin/q$ indicates that the group sizes are equal throughout the layers, and leads to more FLOPs but less weights than the constant number of groups $\rm g = q$.
Since $\rm g$ linearly increases as a function of number of channels, we get relatively dense convolutions at the first layers of the network (large maps,  small $\#$ of channels) and sparser convolutions at the last layers of the network (small maps, large $\#$ of channels).
In these examples, having more parameters at the beginning of the network increases the accuracy, at the expense of more FLOPs. The configuration is advantageous when having a low number of parameters is more crucial than FLOPs. On the other hand, keeping the number of groups constant adds a fixed proportion of parameters and FLOPs to the $1\times1$ convolution, and should be chosen in cases where FLOPs cost as considerably as number of parameters.
As a result, the optimal configuration for an application can be wisely chosen based
on the limitations of the target device.
If there is a constraint on the number of FLOPs, then a constant number of groups can be beneficial, but if the emphasis is on a lower number of parameters, then, a configuration of $\rm g = \cin/q$ will be more suitable for the application.
In addition, the table shows that by a small addition of parameters to the lean network yields higher accuracy, which gets closer to the considerably larger fully-coupled network.

\subsection{Semantic Segmentation} We demonstrate the effectiveness of our proposed network for the semantic segmentation task, which, for instance, can be used in autonomous vehicles, which require real-time inference and by design have less computational power. We adopt two popular segmentation architectures, namely U-net \cite{ronneberger2015u} and DeepLabV3 \cite{chen2017rethinking}. Specifically, we use the general U-net architecture built on top of ResNet as a backbone. That is, we adopt an encoder-decoder scheme, where the encoder is of a standard ResNet architecture and the decoder is based on upscaling operations and transposed convolutions within a ResNet block. Similarly to the classification task, the U-net based on ResNet is used as a baseline. With these settings, we use the baseline with similar networks incorporating various backbones as encoders: MobileNetV2, ShuffleNetV2, ShiftResNet, and ours.
As part of the decoders, we perform convolutions to decrease the number of channels and then perform upsampling, such that in the last layer we have an image, the same size of the labeled image.
For the second batch of experiments, we adopt DeepLabV3's ASPP module to be our decoder, and compare the various encoders. 
We test the networks using two popular datasets - Cityscapes (fine annotated) and PASCAL VOC 2012.
Cityscapes \cite{cordts2016cityscapes} contains 5000 finely-annotated images with 19 categories ranging from road, vehicles, trees and humans. We use the standard train-validation data split as in \cite{cordts2016cityscapes} , i.e.; 2975 and 500 for training and validation, respectively.
We resize the images from $1024 \times 2048$ to $512 \times 1024$ due to memory and computational limitations.
As shown in \cite{pohlen2017full} the reduction in performance is only marginal when down-sampling the images this way.
In addition, we use standard augmentations like random horizontal flips and random rotation of 10 degrees.
The PASCAL VOC 2012 dataset contains 1,464 training, 1,449 validation, and 1,456 test images over 21 object classes (including a background class). We follow standard usage of the augmented PASCAL dataset \cite{hariharan2011semantic} which brings the training set to 10,582 images.
In the training process, we use the ADAM \cite{kingma2014adam} optimizer with a minibatch size of 8 and weight decay of 0.01.
The initial learning rate is 1e-4 and we employ an adaptive learning rate reduction, where upon stagnation of the mIoU metric for more than 5 epochs, the learning rate is decreased by a factor of 10. We use the Focal loss \cite{lin2017focal} as it penalizes wrong segmentations more than correct ones, relative to Cross-Entropy loss. In table \ref{tab:configSegmentation} we summarize the configurations used for the segmentation experiments, where again, we tried to configure the sizes of all the compact architectures to have similar number of parameters and FLOPs.

\begin{figure}[ht!]
	\begin{center}
		\includegraphics[width=.48\textwidth]{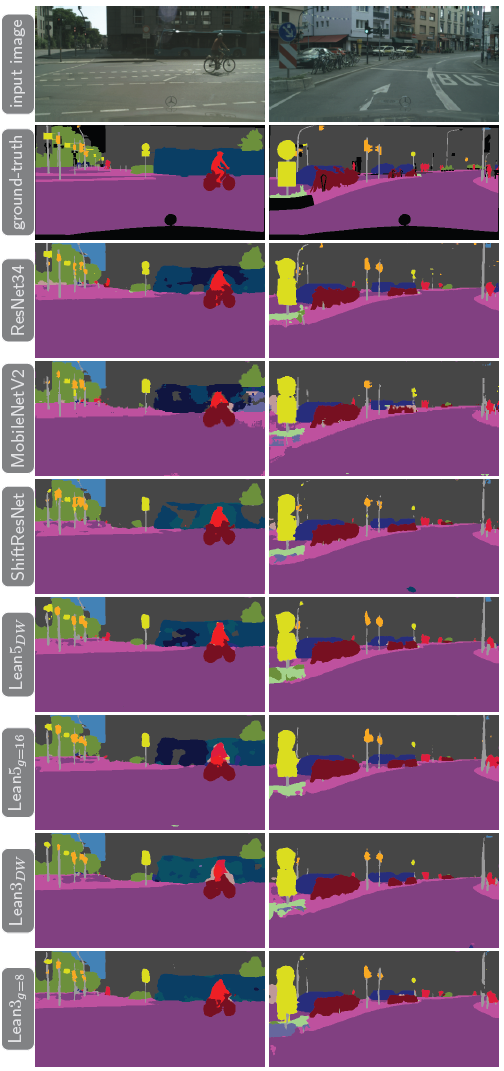}
	\end{center}
    \caption{Visualization of the semantic segmentation results of different networks for two images from the Cityscapes dataset. }
    \label{fig:seg}
\end{figure}

Table \ref{tab:Segmentation} shows the segmentation results. Similarly to the classification results, the lean networks yield performance that is comparable to the other compact architectures. In particular, the grouped lean versions again yield the best accuracy among compact networks, with a slight increase in the parameters and FLOPs. Table \ref{tab:iuPerClassCityScapes} shows the segmentation accuracy per class, and Fig. \ref{fig:seg} shows two example images from the data set and their segmentation result using different networks. For the U-net architecture, the encoder networks are trained from scratch to give equal starting point to all experiments, employing the same training scheme. The networks which adopt the ASPP module as decoder were trained starting from pre-trained models on ImageNet \cite{ImageNet}.

\begin{table}
\centering
\begin{adjustbox}{width=0.5\textwidth}
\small
\begin{tabular}{|c|c|c|c|}
\hline
Type & $\#$ Channels & $\#$ Steps & Strides\\
\hline
MobileNetV2 &32-64-128-256 & 1-2-3-2 & 1-2-2-2\\
ShuffleNetV2 &116-232-464-512 &7-10-10-1 & 1-2-2-2-2\\
ShiftResNet & 64-128-256-320 & 3-4-6-4 & 1-2-2-2\\
(Lean)ResNet34 & 48-96-192-384 & 3-4-6-4& 1-2-2-2\\
\hline
\hline
MobileNetV2 & 16-24-32-64-96-160-320 & 1-2-3-4-3-3-1 & 1-2-2-2-1-1-1 \\
ShuffleNetV2 X1.0 & 24-116-232-464& 1-3-7-3 & 2-2-2-1 \\
LeanResNets & 64-128-256-512 & 3-4-6-3 & 2-2-2-1 \\
ResNet34 & 64-128-256-512 & 3-4-6-3 & 2-2-2-1 \\

\hline
\end{tabular}
  \end{adjustbox}
\caption{Semantic segmentation network configurations. The upper and lower tables refer to the backbones in the U-net and DeepLabV3 architectures, respectively.}
\label{tab:configSegmentation}
\end{table}

\begin{table}
    \centering
    \begin{adjustbox}{width=0.5\textwidth}
    \small
	\begin{tabular}{|l|ccc|}
\hline
U-net's backbone                  & \multicolumn{3}{c|}{Cityscapes} \\
architecture  &  Params[M]$\setminus$FLOPs[B] & Val. acc. & mIoU   \\
\hline
 ResNet34  & 25.95 $\setminus$ 228.4 & 94.1\% &65.1\%    \\
 MobileNetV2   &  3.50 $\setminus$ 31.0 & 92.1\%  & 56.9\%  \\
 ShuffleNetV2    & 3.43 $\setminus$ 36.5 & 90.7\% &  53.5\%   \\
 ShiftResNet       & 3.82 $\setminus$ 48.0 & 93.0\% & 60.0\%   \\
 LeanResNet34 5pt$_{DW}$ &  3.53 $\setminus$ 31.6 & 92.8\%  & 57.9\%  \\
 LeanResNet34 5pt$_{g = 16}$ &4.12 $\setminus$ 36.0 & 92.8\% & 60.2\%  \\
 LeanResNet34 3pt$_{DW}$      &  3.41 $\setminus$ 30.1 & 92.8\%  & 59.2 \%   \\
 LeanResNet34 3pt$_{g = 8}$      &  3.96 $\setminus$ 34.9 & 93.1\%  & 61.7\%    \\
 \hline \hline
 DeepLabV3 backbone                  & \multicolumn{3}{c|}{PASCAL VOC 2012} \\
architecture  &  Params[M]$\setminus$FLOPs[B] & Val. acc. & mIoU   \\
\hline 
 ResNet34  & 25.42 $\setminus$ 39.5 & 94.1\% &73.1\%    \\
 MobileNetV2   &  4.52 $\setminus$ 8.35 & 92.18\%  & 66.9\%  \\
 ShuffleNetV2 [1.0]    & 4.55 $\setminus$ 9.04 & 90.57\% &  65.54\%   \\
 LeanResNet34 5pt$_{DW}$ &  5.12 $\setminus$ 8.17 & 92.48\%  & 67.40\%  \\
 LeanResNet34 5pt$_{g = 8}$ & 6.41 $\setminus$ 9.05 & 93.30\% & 70.31\%  \\
 LeanResNet34 3pt$_{DW}$ & 4.91 $\setminus$ 6.35 &  92.11\% & 66.23\% \\
 \hline
\end{tabular}
    \end{adjustbox}
	\caption{Comparison of our semantic segmentation results with other compact networks. }
	\label{tab:Segmentation}
\end{table}

\begin{table*}
    \centering
    \begin{adjustbox}{width=1\textwidth}
    \small
	\begin{tabular}{|l|ccccccccccccccccccc|l|}
\hline
Method&road&swalk&build. & wall & fence  & pole &tlight & sign & veg. & terrain & sky & person & rider & car & truck & bus & train & mbike & bike & mIoU  \\
\hline
 ResNet34          &  97.2  & 78.3  & 89.1 &  40.5  & 44.2 & 49.8 & 53.2 &  63.9 & 90.0 & 58.0 & 92.4 & 70.9 & 47.1
 & 91.8 & 62.6 & 66.6 & 36.8 & 40.0 & 63.9 & 65.1\\
 MobileNetV2     & 95.5 & 70.4 & 86.4 &  31.9  & 34.3 & 41.4 & 41.3 & 56.3 & 88.4 & 54.3 & 89.6 & 63.3 & 34.8 & 87.8 & 39.4 & 52.1 & 35.0 & 20.9 & 57.7 & 56.9\\
 ShuffleNetV2  & 95.6 & 69.5 & 83.4 &  35.9  & 34.5 & 16.9 & 26.9 & 39.4 & 83.7 & 49.0 & 85.0 & 49.8 & 27.7 & 84.9 & 52.3 & 62.3 & 44.9 & 25.4 & 49.6 & 53.5\\
 ShiftResNet & 96.5 & 74.3 & 87.1 & 38.1 & 37.6 & 43.8 & 43.9 & 55.0 & 88.7 & 54.7 & 92.0 & 64.5 & 39.3 & 88.9 & 49.9 & 62.2 & 32.8 & 31.5 & 59.0 & 60.0  \\
 LeanResNet 5-pt$_{DW}$& 96.6 & 74.5 & 87.0 & 41.5 & 40.7 & 38.0 & 44.3 & 54.0 & 88.2 & 53.5 & 91.0 & 62.6 & 36.7 & 88.8 & 50.7 & 62.4 & 31.0 & 28.0 & 56.1 & 59.2 \\
 LeanResNet 5-pt$_{g = 16}$       &  96.6   &  74.3 & 86.9 & 44.0  & 42.6 & 39.1 & 41.7 & 53.7 & 88.1 & 54.5 & 90.8 & 62.6 & 37.7 & 89.0 & 53.5  & 61.8 & 42.0 & 28.3 & 55.8 & 60.2  \\
 LeanResNet 3-pt$_{DW}$ & 96.6 & 74.5 & 87.0 & 41.5 & 40.7 &  38.0 & 44.3 & 54.0 & 88.2 & 53.5 & 91.0 & 62.6 & 36.7 & 88.8 & 50.7 & 62.4 & 31.0 & 28.0 & 56.1 & 59.2    \\
 LeanResNet 3-pt$_{g = 8}$ & 96.5 & 74.2 & 87.2 & 43.9 & 40.4 &  41.7 & 43.4 & 56.2 & 88.6 & 54.9 & 91.4 & 63.1 & 37.8 & 89.3 & 58.9 & 66.8 & 49.7 & 30.7 & 58.4 & 61.7    \\
\hline
\end{tabular}
\end{adjustbox}
	\caption {Per-class results on Cityscapes validation set using the U-net architecture.}
	\label{tab:iuPerClassCityScapes}
\end{table*}

\begin{table*}
    \centering
    \begin{adjustbox}{width=1\textwidth}
    \small
	\begin{tabular}{|l|ccccccccccccccccccccc|l|}
\hline
Method&background&airplane&bicycle & bird & boat  & bottle &bus & car & cat & chair & cow & table & dog & horse & motorbike & person & plant & sheep & sofa & train & monitor & mIoU  \\
\hline
 ResNet34          &  93.6  & 88.2  & 40.1 &  84.1  & 63.8 & 77.9 & 93.7 &  82.9 & 88.0 & 36.7 & 74.1 & 56.0 & 79.4
 & 77.7 & 81.0 & 84.3 & 56.5 & 80.8 & 42.9 & 82.4 & 70.8 & 73.1 \\
 MobileNetV2     &91.7 & 77.6 & 38.7 &  72.4  & 55.6 &66.3 &85.1 & 79.3 & 80.7 & 32.7& 73.2 &48.9 & 73.8 & 72.0 & 75.8 & 78.1 & 46.6 & 79.4 & 37.9 & 76.7 & 62.8 &  66.9\\
 ShuffleNetV2  & 91.5 & 79.8 & 38.1 &  73.5  & 57.2 & 58.2 & 86.1 & 81.6 & 82.1 & 27.0 & 65.1 & 41.7 & 72.5 &73.0 &76.8 & 79.0 &42.3 & 72.7 & 36.3 & 77.8 & 63.4 & 65.5\\
 
 LeanResNet 5-pt$_{DW}$& 92.0 & 81.3 & 38.3 & 75.5 & 60.2 & 63.7 & 86.0 & 81.3 & 80.9 & 28.4 & 71.9 & 40.7 & 71.7 & 76.3 & 78.6 & 79.6 & 48.8 & 79.4 & 38.5 & 77.5 & 64.6 &  67.4 \\
 LeanResNet 5-pt$_{g = 8}$       &  92.9   &  84.7 & 39.0 & 79.3 & 66.6 & 70.9 &89.8 & 79.6 & 86.3 & 32.3 & 69.7 & 49.2 & 75.0 & 76.4 & 80.9  & 81.9 & 50.4 & 78.2 & 45.0 & 80.6 & 67.8 & 70.3  \\
  LeanResNet 3-pt$_{DW}$       &  91.8   &  78.4 & 36.9 & 76.9 & 62.7 & 66.9 &88.1 & 79.6 & 82.3 & 25.1 & 67.2 & 44.5 & 75.2 & 72.9 & 74.4  & 77.9 & 44.7 & 73.1 & 30.1 & 79.6 & 62.1 & 66.2  \\
\hline
\end{tabular}
\end{adjustbox}
	\caption {Per-class results on PASCAL VOC 2012 validation set using the  DeepLabV3 architecture.}
	\label{tab:iuPerClassPASCAL}
\end{table*}

\subsection{Computational Performance}
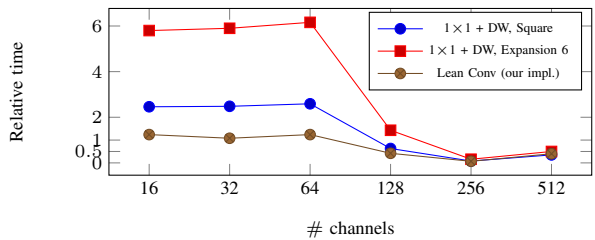
\begin{figure}[ht!]
    \centering
    \scriptsize
\begin{tikzpicture}
     \begin{axis}[width=8cm,height=3.8cm,
     xlabel=$\#$ channels,
     ylabel=Relative time,
     xmode=log,
     log ticks with fixed point,
     xtick={16,32,64,128,256,512},
     ytick={0,0.5,1,2,4,6},
     xticklabels from table={labels.dat}{input},
     ]
 \addplot table {
 16   2.46
 32  2.48
 64  2.59
 128  0.63
 256  0.081
 512 0.35
 };
 \addplot table {
 16  5.8
 32  5.9
 64  6.16
 128 1.43
 256 0.16
 512 0.5
 };
 \addplot table {
 16  1.24
 32  1.08
 64  1.24
 128 0.42
 256 0.07
 512 0.4
 };
\addlegendentry{\tiny{1$\times$1 + DW, Square}}
\addlegendentry{\tiny{1$\times$1 + DW, Expansion 6}}
\addlegendentry{\tiny{Lean Conv (our impl.)}}
 \end{axis}
 \end{tikzpicture}
    \caption{Relative timings of reduced convolutions compared to a $3\times3$ convolution (lower is faster). The expanded and square $1\times1$ convolutions has the same number of weights.
    }
    \label{fig:timing}
\end{figure}

We compare the latency of our CUDA implementation of the lean convolution with two other combination of layers, comprised of a $1\times1$ convolution that is followed by a depth-wise convolution. In one combination we use $c_{in}=\cout$, and in the other $\cin\approx 6\cout$, but with the same number of weights. Such layers are applied in \cite{sandler2018mobilenetv2}. We compare the runtime of a typical network: the first layer consists of 16 channels of $512\times 512$ maps, and the maps are coarsened by a factor of 2 when the channels increase by a factor of 2 (i.e., for $512$ channels the images are of size 16). We use a batch size of 64, and compare the runtime of a NVIDIA GeForce 1080Ti GPU for the task. The implementation for the other convolutions is based on PyTorch's $1\times 1$ and grouped convolutions using CUDA 9.2. Figure \ref{fig:timing} summarizes the results. The depthwise convolutions dominate the low channels layers, while all combination converge to the cost of the $1\times1$ convolution as the channels increase (and the depthwise layer becomes negligible). Our implementation of \eqref{eq:Klean} is comprised of a standard 4-point convolution for each channel followed by a matrix multiplication using {\tt cublas} for the $1\times1$ part, to use the highly optimized {\tt gemm} kernel. Our implementation is faster because the shiftIm2Col approach is not efficient for small group sizes (1 in this case). The clear advantage of the lean operator over the expanded combination is the fewer feature maps that undergo the spatial convolution. Although our implementation applies the $1\times1$ and depthwise convolutions separately for each sample, our experiments show that this yields a performance gain compared to a completely separate multiplication for the whole mini-batch.

\section{Conclusion} \label{sec:discussion}
We present LeanConvNets, a family of efficient CNNs that reduce the number of weights, and floating point operations with minimal loss of accuracy.
LeanConvNets can be obtained from existing CNNs by replacing fully-coupled convolution operators by lean operators that are the sum of grouped and $1\times1$ convolutions.
The group size serves as a hyperparameter that allows the user to trade off computational cost and accuracy.
Additional savings can be realized by the proposed five-point and three-point stencils.
Those savings will be more pronounced for 3D and 4D imaging data.

In our experiments, we apply various configurations of LeanConvNets to image classification and segmentation tasks.
In our tests, the LeanConvNets perform slightly better than other reduced architectures, and is almost as effective as their fully-coupled counterpart.
We also demonstrate in a direct comparison that the addition of depth-wise and $1\times1$ convolution reduces the computational time.

Our future research aims to further optimize implementation of the lean convolutions on GPUs, as well as investigate optimization of such implementation on other devices. In addition, it is worthy to investigate and characterize the hyper-parameter choices of the lean convolution (groups, stencil size, multiplication algorithm), as these choices should be guided by the hardware \cite{marculescu2018hardware}. We also plan to examine the efficiency of the lean operators in challenging 3D applications such as video analysis on limited devices \cite{fassold2015real}, where the small stencil size is more beneficial.

\section*{Acknowledgements}
LR’s work is supported by the US National Science Foundation (NSF) award DMS 1751636.
This research was partially supported by grant no. 2018209 from the United States - Israel Binational Science Foundation (BSF),
Jerusalem, Israel. ME is supported by Kreitman High-tech scholarship.
\ifCLASSOPTIONcaptionsoff
  \newpage
\fi


\bibliographystyle{IEEEtran}
\bibliography{LeanCNN}

%







\end{document}